# AD-EYE:  A Co-simulation Platform for Early Verification of Functional Safety Concepts.

**Authors: Naveen Mohan, Martin Törngren**

**Corresponding Author email: naveenm@kth.se**

## Abstract

Automated Driving is revolutionizing many of the traditional ways of operation in the automotive industry. The impact on safety engineering of automotive functions is arguably one of the most important changes. There has been a need to re-think the impact of the partial or complete absence of the human driver (in terms of a supervisory entity) in not only newly developed functions but also in the qualification of the use of legacy functions in new contexts. The scope of the variety of scenarios that a vehicle may encounter even within a constrained Operational Design Domain, and the highly dynamic nature of Automated Driving, mean that new methods such as simulation can greatly aid the process of safety engineering.

This paper discusses the need for early verification of the Functional Safety Concepts (FSCs), details the information typically available at this stage in the product lifecycle, and proposes a co-simulation platform named AD-EYE designed for exploiting the possibilities in an industrial context by evaluating design decisions and refining Functional Safety Requirements based on a reusable scenario database.

Leveraging our prior experiences in developing FSCs for Automated Driving functions, and the preliminary implementation of co-simulation platform, we demonstrate the advantages and identify the limitations of using simulations for refinement and early FSC verification using examples of types of requirements that could benefit from our methodology.

## I. Introduction

While automated vehicles promise huge benefits to society, they are notoriously hard to verify as safe. A fully automated vehicle essentially operates in a relatively unconstrained environment with little or no human help in making life-critical decisions. The increase in software and system complexity required for high levels of automation is unprecedented. Some of the tasks, in particular perception, have required adoption of several machine learning and AI techniques including Deep Neural Networks (DNNs).

The use of DNNs and other opaque non-deterministic techniques dictate the use of a black box approach towards testing components to ensure safety of the automated vehicle. Some estimates such as in [1] and [2] suggest the need of billions of kilometres of driving before the vehicles can be certified as safe. Since this is beyond current capabilities of most OEMs to perform with real vehicles on road, focus has shifted into simulation and the ability to perform most of the kilometres in a virtual world rather than the real one. It is generally accepted today that simulation-based testing is inevitable for Automated Driving verification.

This paper aims to incorporate the use of simulation-based testing for the purposes of early stage requirement generation and refinement, using a common validation strategy to assist disparate development groups in working together. In particular, this paper aims to use simulations to aid design decisions made by architects (and to quantify them) for the work product Functional Safety Concept (FSC) as required by the concept phase of ISO 26262 [3].

The authors believe that there is a fundamental difference in the primarily top-down approach prescribed by ISO 26262, and the bottom-up component-based thinking that is prevalent in the automotive industry today. The authors see simulations as an invaluable tool in design which can be used to unify the top-down and bottom-up approaches, with the use of progressively more accurate simulations through the product lifecycle.

Leveraging our experiences from various Swedish national and European projects with the automotive industry, we have designed a platform to enable our research group to collaborate on common use cases within Automated Driving while still being able to focus on their narrower area of expertise. The configurable platform, named AD-EYE serves also as an educational tool to enable students to gain practical experience in working with Automated Driving systems.

The rest of this paper is divided into the following sections: Section II gives a short background and introduces terms used through the paper. Section III then describes the problem and ends with the requirements for the platform. The design choices related to fulfilling the requirements are discussed in Section IV, including details of the technology used, trade-offs etc. Section V describes the current status of the platform, some related work and discusses the known limitations. We end with the conclusions and future work in Section VI.

## II. Background

### *Terminology*

For the terms **Scene and Scenario**, we use the definitions by Ulbrich et al. in [4] *"A scene describes a snapshot of the environment including the scenery and dynamic elements, as well as all actors' and observers' self-representations, and the relationships among those entities"*. A scenario is defined in relations to scenes by Ulbrich et al. in the same source as: *"A scenario describes the temporal development between several scenes in a sequence of scenes. Every scenario starts with an initial scene. Actions & events as well as goals & values may be specified to characterize this temporal development in a scenario."*

Terms related to Automated Driving such **as Levels of Automation, Operational Design Domain(ODD)**, are used in the context of SAE J3016 [5]. Using the definitions by Ulbrich et al., we extend the definition of the ODD to be a set of *Scenarios* that *the Automated Driving Feature* should be able to handle as part of its nominal functionality.

**ADI: Autonomous Driving Intelligence**. A term from [6] used to refer to the additions to the legacy platform required to enable Automated Driving, roughly corresponding to the role a human driver plays in a non-automated vehicle using the 'Observe, Orient, Decide and Act loop [7] as a reference. Used to distinguish the additions to the platform, from the entire functionality for automation (represented by the related term Automated Driving System in [5]). The term ADI is used for only for Automated Driving Features that are SAE L3 and above.

The ADI may comprise of one or more independent control **Channels**, each with capacity to control the vehicle albeit with different goals. For example, in the case study described in Section IVc the ADI is designed with two channels, the goal of the Nominal Channel corresponds to the mission of the vehicle itself and it is responsible for the **Dynamic Driving Task (DDT)**. The Safety Channel operates with a much smaller time horizon with a goal of bringing a vehicle to a stop as safely as possible. Channels may be aware of each other and may actively handover control to each other e.g. when errors are detected.

Terms related to safety engineering such as **safety, functional safety, hazards, Automotive Safety Integrity Levels (ASILs) Functional Safety Requirements (FSRs) etc.** used in this paper is are used in the context of ISO 26262, unless otherwise mentioned.

### *Safety and Architecture in ISO 26262*

Safety and architecture are intertwined within ISO 26262. Through each phase of development, as safety requirements get refined from Safety Goals (SGs) to Functional Safety Requirements (FSRs) to Technical Safety Requirements (TSRs), these requirements are allocated to architectural elements available at that level of abstraction. The architectural abstraction in the concept phase of development termed the Preliminary Architectural Assumptions (PAA) will be the focus of this paper and the term *elements* refers to the architectural elements that are part of the PAA. The FSC consists of the FSRs allocated to elements of the PAA in conjunction with the reasoning as to how they serve in mitigating hazards, and evidence of this reasoning.

The elements of the PAA are essential for the purposes of a technique commonly used within the industry, the so-called *ASIL decomposition,* where the ASILs assigned to safety requirements can be effectively be reduced by duplication of the requirements, and allocation to *sufficiently independent architectural elements*. ISO 26262 does not specify the exact nature of the PAA or its elements and mentions it to be obtained from an external source. In this paper, we will use the definitions and the data model provided in [8].

## III. Problem Description

The essence of an FSC is the allocation of FSRs to the elements of the PAA, along with reasoning about how the concept fulfils the SGs assigned to the system. ISO 26262 allows for the tailoring of FSRs to better match the PAAs including lowering of the ASIL levels using architectural redundancies if the SGs are not violated. The Mechatronics and Embedded Control Systems Division at KTH is and has been part of several projects relating to Automated Driving over a period of time, collaborating with several European universities and OEMs. As such, over time, we have had access and the privilege to contribute to several FSCs for Automated Driving Features. We have experienced that while the tailoring of FSRs to match the PAA elements is straightforward to do in theory, it is not in practice.

The increasing emphasis in top-down design, that ISO 26262 brings, typically works in parallel to the bottom-up approach in the industry as sensor choices, etc. are influenced also by non-technical factors such as choosing strategic relationships, and building around existing limitations. The reader is directed to [8] for an extended description about the challenges in design of FSCs for high levels of automation. This paper will discuss only a subset of the challenges.

While higher levels of Automated Driving have not been implemented in series produced vehicles by any OEM at the time of writing, many of the key components such as lane keeping, automatic emergency braking etc. are now part of many vehicles, along with the sensors needed to achieve them. One of the most important features of early concept design for generating the PAA is the prevalence of legacy systems, and the gains to be made by reuse over new development within the cost-sensitive industry. Modern vehicles are seldom built completely anew and are instead built from a platform of components in compliance with a product line. The reuse of known components and features allows for simplified verification and lower costs, but the reuse of legacy elements impose constraints on the rest of the system [9].

To take a simplified example, if the ADI were to be built for an L4 function: An option *could* be to reuse the forward-facing camera already present in platform introduced for the pre-existing *Advanced Emergency Braking* feature before. Reusing the camera module would however also mean inheriting constraints such as the field of view of the camera designed for the older feature. The ADI would likely need other sensors to complement its field of view and it would be beneficial to optimize the choice w.r.t. the sensors already in the platform. Equally, another option *could* be to ignore the current sensor configurations available in the platform and develop new sensor solutions for the new feature. The design decision that chooses between the two options, influences the capabilities of the sensors and the structuring of the FSRs.

A way to proceed with the design decision faced in the example, is to have a metrics-based evaluation of the options to ensure complete coverage of the desired ODD, and then choose based on the scoring on the metrics. To do so, in the case of Automated Driving, requires detailed information that may not yet be available e.g. Even a simple scenario of a vehicle X overtaking on the left of the ego vehicle, could have drastically different outcomes based on the behaviour of the X, environmental conditions, the perception and the motion planning algorithms of the ego vehicle. This dynamic interplay is difficult to test on the road and is arguably harder to define in terms of FSRs at design time without insight into the underlying technology that will be used in latter stages of development.

The information needed may be too detailed to be in the competence sphere of the architect who must rely on different Domain Experts for an answer to proceed with his work. A compounding problem is that even the Domain Experts may not have accurate answers in the early stages of development. E.g. a perception algorithm may be rated w.r.t. a given data set (as is common with machine learning and AI techniques), but the exact performance of an algorithm depends on its use case i.e. the ODD. Estimates are inaccurate in the early stages of development when all the scenarios that make up the ODD are not known. The ODD is constantly refined during development as testing reveals edge cases, accident data, and failure rates from the field may lead to multiple revisits of the design.

The architecting team must rely on several Domain Experts and their specialized tools, constantly weighing answers from one in relation to the others so as to clear out inconsistencies caused by the early stage of development. E.g. Is the perception field of view around the vehicle still enough with the new estimates for braking distances from the vehicle dynamics division? While this has worked in the past, with the scale of Automated Driving and the different types of expertise needed, keeping early estimates of answers consistent is a challenge.

This paper advocates the use of a simulation platform that unifies the early concept design under the blanket of simulation-based validation. i.e. having the ODD formalized into a set of scenarios, and metrics to judge how well a given concept performs. By having the same set of scenarios for the decoupled parts of the organization, the impact of changes brought in due to new information on an existing concept can quickly be analysed and appropriate changes can be made. The ODD is used as a validation target through the development process to ensure that development teams across the organization meet the current known goals in the right context.

This paper further discusses the development of a simulation platform that makes use of a separation between the ADI, the base platform and the environment, enabling the management of uncertain information that is subject to change through the product development lifecycle. The platform named AD-EYE enables the use of legacy as well as new code, multiple control channels to use the code, configurability with different specialized tools and several ways to create and maintain the ODD to address the concerns mentioned. The choice of keeping the scenarios separate from the control channels is logical as they are the final validation target. This choice further allows for the independent development of different aspects of the control channels such as clustering, motion planning etc., within the same framework.

The focus of this paper is to detail the ideas behind the design and development of a co-simulation platform for the FSC design that can be reused through different development phases. As such, it does not aim to explain an entire safety assurance strategy, nor important subtopics of functional safety such as tool qualification which are highly case-specific.

The requirements listed below, developed based on our analysis of the problem, our experience within the industry, and several FSCs that we have had access to, were used to develop the platform.

### *Simulation Platform Requirements:*

R1. The platform shall support multiple sources of input data to allow for ease of world creation and scenario generation.
R2. The platform must support test automation across multiple created worlds and scenarios.
R3. Scripting of the test automation process must be possible using standard scripting languages.
R4. The platform shall use open source code for control and decision logic as far as possible.
R5. The platform should provide a common base functionality with a configurable modular design.
R6. The platform shall support physics-based worlds.
R7. The platform shall support modifiable sensor models.
R8. The platform must provide support for the most common sensors within the automotive domain. The platform must provide support for Radar, Lidar, GPS, IMU, Camera and Ultrasonic sensors.
R9. The platform must provide ground truth data during the running of the simulation.
R10. The platform must provide the ability to allow a control channel to control the simulated ego vehicle.
R11. The platform must provide the ability to control multiple actors i.e. vehicles other than the ego vehicle or pedestrians
R12. The platform must support the ability to run and evaluate multiple control channels simultaneously.
R13. The platform must provide the ability to prioritize control signals between the different channels.
R14. Each control channel shall provide control signals that complies to the same specification.
R15. The platform must be able to distinguish between the control signals from different channels.
R16. The platform must be able to distinguish multiple sensors of the same type.
R17. The platform shall support creation of handcrafted worlds and scenarios, and subsequent adaptation of these.
R18. The scenario database shall be reusable in that it shall be independent of utilized perception and control logic.
R19. The linking of sensors to control channel must be configurable to adapt to the needs of each channel.
R20. The platform must possess the capability to inject faults during runtime.
R21. The platform must provide the ability to script fault injection in the same interface as test automation.
R22. The platform must support co-simulation standards such as FMI for extended and specialized simulations.
R23. The portability of the code generated from the platform must not be limited to any particular hardware configuration.

## IV. Design choices

Several design choices have already been made in the development of AD-EYE. This section discusses the choices made, their relation to the requirements presented before and the trade-offs made. To explain the design choices made in context, subsection a. and b. detail the technical design choices made and the reasons behind them, and subsection c. explains the choices in context using a case study.

### *a. Software*

In the current state of the project, some of the most critical design choices have been made. The Robotics Operating System (ROS) has been chosen as the underlying middleware due to its ubiquity in the robotics domain, the possibilities for reuse with the large number of off-the-shelf-packages enabled by its distributed architecture, and the visualization it provides.

For the world and Scenario modelling, the physics-based simulation tool PreScan [10] has been selected. This tool provides test automation capabilities, physics-based worlds and sensor-based interpretations of these worlds. Alternative open options such as Gazebo were considered and discarded due to its limitations in the fidelity of the world modelling as discussed in e.g. [11]. The fidelity of the modelling is a strength in PreScan enabling relatively sophisticated sensor modelling and usage. PreScan while close-sourced, gives access via standardized interfaces, possibilities to import data from sources of interest such as the German in Depth Accident Study (GiDAS) [12] which is arguably the largest database of its kind, Open Street Maps which can be used to recreate real-world locations etc.

Following common automotive praxis, where localization within a known map is preferred over SLAM (Simultaneous Localization And Mapping) algorithms, AD-EYE allows for the automated generation of highly detailed maps of any arbitrary world. This feature developed within the platform, allows for extraction of data from the simulation environment for use by the ADI. In the workflow, this is seen as a separate mapping step performed in ideal conditions and the simulation equivalent of HD-Maps.

In essence, AD-EYE is a co-simulation platform that requires both Simulink and ROS to be synchronized in time. To this end, all nodes written in ROS used the *ROS::time* API to allow for transition between wall-clock and simulated time. A node that publishes the current simulation time on the required ROS topic was created to be used within the Simulink models. Simulink is the master in this co-simulation platform, maintaining control of simulation time, combinations of variables for combinatorial testing, performing automated tests and provides fault injection capabilities for the sensors and vehicle dynamics. Simulink is also used for controlling fault-injection inside the ADI control channels by interfacing with the channels. Simulink subsystems used for running the entire simulation i.e. starting, configuring and killing of the ROS framework core running on Linux computer, also needed to be developed. Thus, AD-EYE can exploit the availability of the large ROS code base at design time and control a chosen instance of it over runtime.

All the ROS interfacing code within Simulink utilizes blocks from the MathWorks' Robotics System Toolbox included with MATLAB 2017B. At the time of writing, this has led to some bottlenecks in performance particularly with the processing of large sets of data such as point clouds. There are thus performance gains still to be made by moving to compiled S-functions that utilize lower level coding languages i.e. C natively.

Further adaptations needed to be made to harmonize the simulation particularly with regards to transforms between the different frames of reference within the simulation. The ROS package *tf2* was extensively utilized for this purpose to let the ADI understand the relations between frames in native ROS formats. Simulink subsystems were created to calculate and publish the various required transformations between the frames to the ROS network.

The open-source Automated Driving stack Autoware [13] has been selected as the Nominal Channel within the AD-EYE platform because it provides a full configurable stack of base functionality where all modules have been tested to be interoperable. Autoware has been successfully used in several projects with both real and virtual vehicles around the world. For AD-EYE, Autoware acts as a modifiable base where only the parts of the code that are required by a specific sub-project needs to be modified, empowering researchers to focus on only their topic of interest e.g. a specific localization algorithm etc. while still being able to work in the same framework and context as others.

The integration with Autoware is seen more as a metric for the compatibility with ROS than an absolute necessity for the platform. From the AD-EYE perspective, Autoware is only a single control channel and developers can still reuse the interfacing code with ROS, and the rest of the features of AD-EYE to build their own control channels. The link between ROS and Simulink i.e. between the ADI and the world is only standardized sensor input and control output interfaces. AD-EYE does not inherently place any limits on the number of control channels using the interface and the sources of this information, i.e. sensor based (real/simulated) or ground truth, is transparent to the ADI. The platform uses the well-known pipes-and-filters architectural pattern to ensure that the interfaces between the ADI and the world are only as per common sensor inputs. The strict interfaces between the ADI and the world allows for portability of the ADI code (with relatively minor modifications) to other platforms such as ECUs like the Nvidia Drive PX2 [14].

Faults can be injected into the system directly from Simulink (for sensor/vehicle dynamics related faults) or using the Simulink interface to ROS (for faults in the control channels). For example, the envisioned method of fault injection for sensor faults, involves the creation of subsystems in Simulink per sensor, linked to a fault model. This facilitates in the case of the camera, pixel level manipulation on cameras, and faults such as dead pixels, or gaussian noise over a range of pixels, can easily be simulated giving the ADI "accurate" faulty input. On the ADI/ROS side, topic remapping helps extract information from a module/unit level, and the open-source nature of the code enables the awareness of the internal states of the ADI at runtime using hooks allowing for monitoring and manipulation. There are thus vast possibilities of injecting faults in code of interest, monitoring the results in the internal states of the ADI and the entire system and reacting to the dynamic changes in real-time.

Thus, the design decisions made, enable the users of the AD-EYE platform to create new/import and modify existing arbitrary worlds, design scenarios, specify dynamic behaviours of the vehicle and algorithms for the ADI, and automate several tests under varying conditions.

### *b. Hardware*

Two hardware configurations have been chosen for utilizing the software setup.

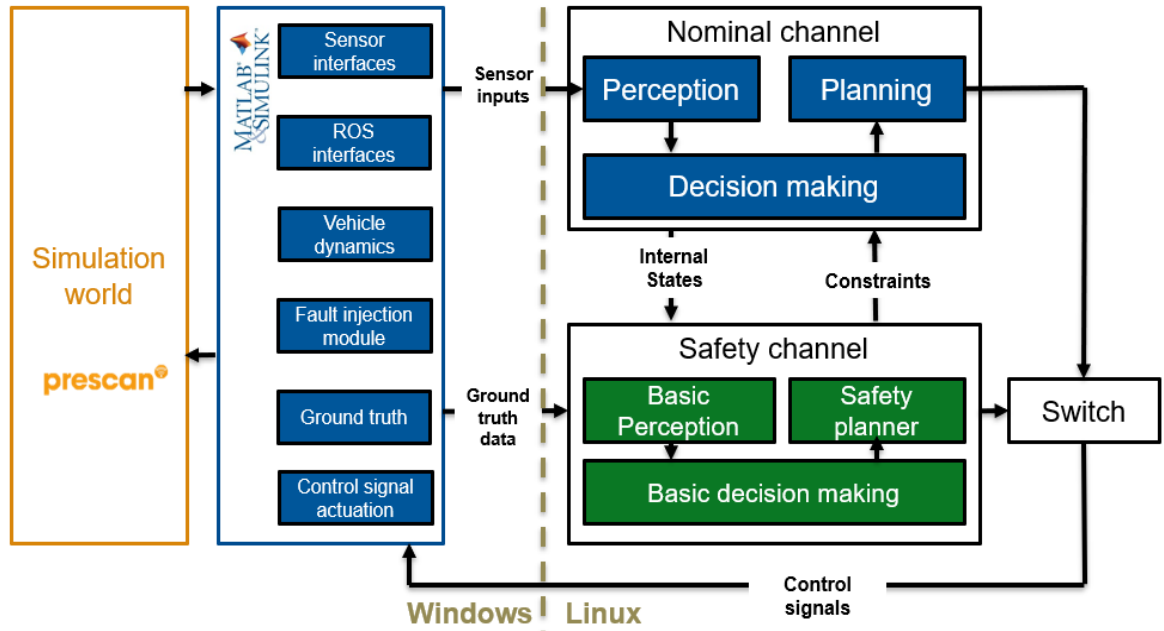

*Figure 1: Case study architecture*

For the purposes of development, where we foresee the use of many iterations and small changes in the design of the environment, *Config 1* comprising a single x86 PC will be used. The x86 PC has an intel 8700k processor, 2 Nvidia 1080Ti GPUs and 64 GB of RAM. Natively this PC runs the Ubuntu distribution of Linux with a thin hypervisor layer utilizing the KVM/QEMU libvirt packages. Half of the available resources including CPU cores, GPUs and RAM are allocated to the virtual machine. The choice of the virtualization layer allows for near native usage of the GPU resources using a PCIe *passthrough* of the graphics card [15]. The guest Virtual Machine has windows as the Operating System and runs PreScan over the passed-through GPU, while the host machine runs the ADI over Linux. Config 1 allows a single developer to simultaneously develop on both the Windows and Linux environments speeding up the initial iterations.

*Config 2* comprises a x86 PC running windows natively for PreScan, in conjunction with the Nvidia Drive PX2 running Autoware. This configuration allows for the code run on Config 1 to be tested on realistic hardware while freeing up the GPUs on the x86 PC to be utilized for the resource intensive physics-based GPU usage for PreScan's world modelling. Config 2 is intended for overnight automated running of simulations.

A potential, yet unexplored midway solution between configs 1 and 2 is to generate code from Simulink and run the entire collected code base on the Linux system. This option compromises the advantage of running on realistic hardware in Config 2, in lieu of the simplification of hardware platform and the speed gained by the reducing the Simulink engine's overhead.

### c. Case Study

As an example we instantiate the architecture proposed in [16] to demonstrate the capabilities of, and to evaluate our platform. Figure 1 shows an abstracted version of the architecture with two desired control channels viz. 1. the Nominal Channel and 2. The Safety Channel. The world modelling tool PreScan and Simulink are tightly coupled and once the experiment is designed, Simulink serves as the only interface to PreScan at run-time. The directed arrows all denote one or more ROS topics for communication. The dashed line in the centre of Figure 1 denotes the split between the code base that is Linux based i.e. the ADI code (on the right of the line) and the code that can be run on either windows or Linux i.e. the world & scenario modelling (on the left of the line) depending on the hardware configuration used.

The Nominal Channel is assumed to be the one primarily in control of the vehicle within the ODD, with the Safety Channel assuming control only in cases such as failure in nominal functionality etc. The assumption the authors make in [16] is that the Nominal Channel is a more complex piece of software that is more likely to fail, while the Safety Channel will exhibit only the most basic functionality needed to take the vehicle to a safe stop. Arbitration between these channels of control is fixed in that the Safety Channel can always override the Nominal Channel.

Also implicit in the assumptions, is that the sensory input for the safe channel will be provided by sensors that are classified to be robust within the ODD. In line with these assumptions, the Safety Channel within this case study is provided from ground truth data from Simulink, and the Nominal Channel is provided data generated from the interpretation of the simulated world by physics-based sensor and world models. Thus, the Nominal Channel alone is subject to real-world properties such as sensor occlusion, energy absorption, direction of light source, weather conditions etc., while the Safety Channel acts as an oracle for testing. The Safety Channel decides independently of the Nominal Channel if the vehicle behaviour can cause an unsafe situation, and disables the Nominal Channel while simultaneously triggering a safe stop using the planner described in [17]. The Safety Channel essentially monitors the Nominal Channel and the environment using ground truth data, and disconnects the Nominal Channel as needed in case of faults or undesirable behaviour within the Nominal Channel using the Switch component in Figure 1.

Using this setup, and given an ODD as a set of scenarios, AD-EYE provides a capability to automate, and run the scenarios under all combinations of the variables that are of interest. Examples of variables could be start position of different actors, sensor mounting points, initial speeds, lighting conditions etc.

While this does not mean that an equivalent performance will be obtained from the perception and decision-making stack once a transition is made to real-world driving, it helps refine safety requirements that stem from architectural design decisions made early in the development phases. E.g. For a given vehicle dynamics profile, how many lidars of type X are needed to complement the existing legacy cameras of type Y for Automated Driving feature Z, where should they be mounted in conjunction to existing sensors? Or what are the operational limits in which a Nominal Channel can work while still maintaining the safety constraints derived from the hazard analysis? etc. Typically, the safety constraints available from traditional safety analysis techniques are based on assumptions on system behaviour at design time and have large safety margins. Simulation techniques can be used to explore these margins with progressively more detailed (as the project matures) behaviour from the control channels and help in generating test cases for edge scenarios to be performed real vehicles.

We have used the platform to test some of the functional requirements derived from the PAA described in our work in [18] and refine them. In addition, it was possible to validate the assumptions behind some design decisions such as the needed field of view of certain sensors. However due to the limited maturity of the platform at this stage, we have not been able to work through all the possible functional requirements. This is seen as future work. What our limited study helped with concretely, is showing that the AD-EYE platform was able to scale to multiple channels of control and helping us decide metrics that should be used to judge the safety of a particular scenario. The study as also helped us prioritize between different features being developed for the platform.

## V.     Current Status and Discussion

While most of the ideas discussed in this paper have their basis in existing state of the art or state of practice, we have not found another paper that discusses the challenges of designing the PAA and the FSC in an automotive context that provides a tangible solution. Publications such as [19] emphasize the need for early evaluation of safety but do not go into the details of what can be achieved in the early stages. Others discuss the importance of early reduction of risk [20], the use of scenario databases has been discussed in other projects such as the European project PEGASUS [21], [22] but not in the context of architecting.

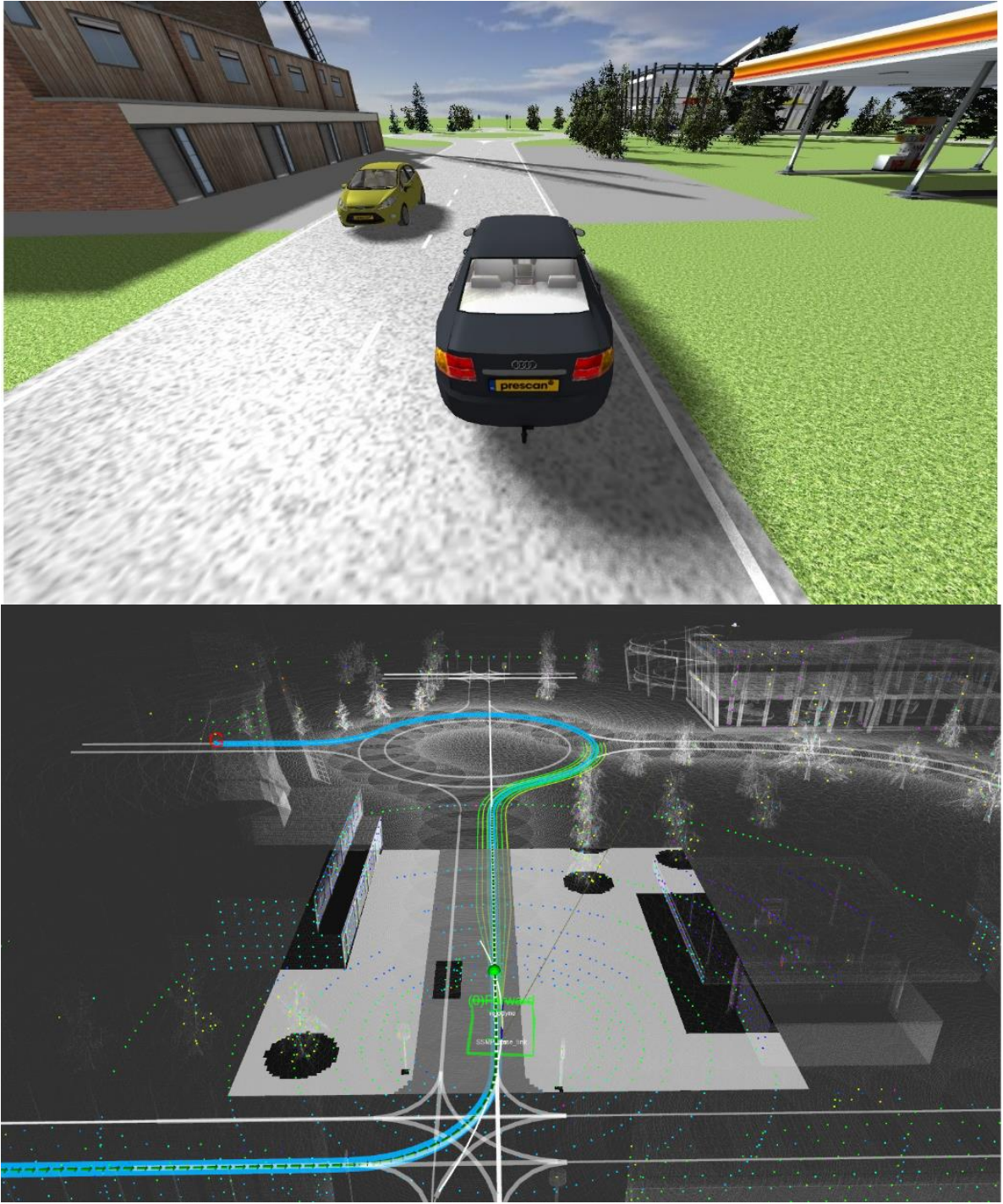

*Figure 2: A hand-crafted simulated world (top) and the internal representation of the world in the ADI during simulation. (bottom)*

The need for simulation is well known within the industry. Waymo and General Motors Inc. who are arguably the leaders in the industry have released safety reports [23] and [24] compliant to the NHTSA advisory [25], which outline their use of simulation in their systems. The utility of simulations is also evident in the scale of verification that can simply not be achieved using real-world driving. Waymo, for example, drives 8 million miles in simulation every single day while it has only been able to accrue a cumulative of 4-5 million real world miles over all of its history [24] [26]. However, the details of how they use their simulations to make their architectural design decisions is not disclosed in their published documents. Furthermore, the documents do not give guidance on how such a platform could be developed.

Other simulation platforms that are seen in literature such as [27] which discusses a co-simulation using MATLAB and ROS usually miss out on addressing the challenges of defining the ODD, leveraging test automation, and the complexities of world modelling with varied data input sources, unlike AD-EYE.

The platform has reached its first milestone in the capability to control a vehicle within PreScan using the ADI channels i.e. the Nominal and Safety Channel, and established interactions with the Safety Channel. Figure 2 comprises a pair of screenshots from a demonstrator experiment where an arbitrary world in PreScan created, and its equivalent interpretation from the perspective of the ADI. The figure highlights several important features from the platform. The top half of Figure 2 is the view, during run-time, of a world created using a GUI at design-time. AD-EYE features were then used to script the mapping of the environment and provide it to the control channels for use during simulation.

The bottom half of Figure 2 shows the two control channels of the ADI in action trying to control the vehicle (green box) in its mission of reaching the goal pose (marked with a red G). The white dots represent the pre-loaded maps and the colours represent what is currently visible to the 360-degree lidar sensor that is mounted on the vehicle in PreScan. The quality of the fidelity of the mapping phase can be seen in that the shapes of trees etc. are clearly visible. Keeping the colour scheme where white represents preloaded information, the white lines represent traversable lanes using the Autoware lane representation format. The coloured lines that traverse through the vehicle represent different possible trajectories that may be by the current desirability (indicated by different shades of colour). The selected trajectory to follow as chosen by the ADI's Nominal Channel is coloured in blue.

The Safety Channel's view of the world is more simplistic and uses the Grid_Map package from ROS to save ground truth information in cells that represent world locations. The information relevant to the safety planner i.e. about the area closest to the vehicle is converted according to the format needed. This area is represented as a solid box in the lower half of Figure 2 and is assigned a safety rating for each cell. A darker colour in the box e.g. the cells of the oncoming vehicle, the buildings or trees indicate an unsafe location.

It is possible today within AD-EYE to change several variables such as environmental conditions e.g. weather, road friction etc. and automate tests to analyse (and influence) how internal states within the system reflect these changes. It is also possible for the Safety Channel to step in and activate a safe stop manoeuvre if it deems that the Nominal Channel has encountered an error or unsafe behaviour.

We have thus made good progress towards meeting the simulation platform requirements detailed in the previous section and so far, our design choices seem reliable in that we now have a configurable platform in controlling a virtual vehicle. We also have gained access to many import sources for obtaining new worlds in simulation and the ability to create custom scenarios as needed. A limitation in PreScan as a choice is the somewhat limited vehicle dynamics options w.r.t. configurability and the need to use other tools for simulation of vehicle dynamics. PreScan provides several different ways to integrate different tools with itself such using Functional Mockup Interfaces [28]. The use of PreScan thus also helps us having a skeleton framework where more specialized tools can be connected as needed.

Some of the compromises made during the development have been the choice of ROS where we gained access to a large flexible code base but lost the ability to simulate some characteristic embedded systems behaviours such as scheduling, real-time guarantees etc. This is one of the limitations of choosing a middleware solution that is designed to work primarily with soft-real time systems at best. Another disadvantage with the selection of ROS was that it has a single point of failure in the *ROS Master*, a process needed for ROS nodes to coordinate communication. This makes ROS in its current form unrealistic for use in safety-critical systems in production, but it does allow for fast prototyping and we capitalize on the fact that for design decisions early in the product lifecycle e.g. in the concept phase of ISO 26262, the choice of technology is not as important as the functionality itself i.e. testing/manipulation of functional behaviours by virtue of standardized interfaces, modularity and open source code.

Another compromise is the fidelity of the world models used within PreScan. While realistic, they do not eschew the need for in vehicle testing. The use of standardized interfaces allows for changes later as needed. With the many compromises made, the AD-EYE platform might not be the most optimal solution for FSC design, but it is achievable at reasonable costs and its modular design helps for expansion as needed.

Several of the challenges described in section III for the industry are also applicable to academic environments. The issue of distributed development for example also occurs within research where different research groups may focus on one aspect of Automated Driving in lieu of others. Many universities also face a manning issue in that they cannot compete with the budgets and manpower of OEMs. AD-EYE was achieved with academic research funding and presented to the industry as a full solution. In addition, AD-EYE is proving to be a very pedagogic learning tool giving students a visual way of learning about the different aspects of Automated Driving and their interrelationships.

The AD-EYE platform at the time of writing, functions as it should, but suffers from some performance bottlenecks. Ongoing sub-projects for fixing these have been started including the porting of the ADI code into platforms such as the Drive PX2, improvements to the Simulink-ROS interfaces, improvements to the test automation framework etc. and better performance is soon to be expected.

# VI. Conclusions and Future work

In this paper, we have analysed the information typically present with an organization at the concept phase within the product development lifecycle, derived requirements for a test platform, and provided a case study where we evaluated the platform with an FSC.

The results, though preliminary, are satisfactory and give credence to some of the design choices that we have made. By our evaluation and use of the platform, we have been able to perform and prioritize amongst the features of the platform still under development.

We have further, already used AD-EYE in an academic context with students and found it to be an effective learning tool, giving students a jumpstart into the field of Automated Driving with a hands-on experience. Currently about 9 Master's students and 3 full time employees have worked on projects utilizing the platform in about 4 different sub-projects, with 2 more sub-projects scheduled to begin under new students soon. While these projects have goals of their own, each contribute to the core framework that AD-EYE provides improving the performance of the platform. Results from these projects that are being generated such as quantitative metrics for evaluation and comparison of FSCs, the test of algorithms developed on the AD-EYE platform in KTH's research prototyping vehicle the RCV [29], are the subject of future publications.

We believe that others can benefit from the design choices in this platform, as presented in this paper. While the code for the Nominal Channel used in the case study i.e. Autoware is already open source, the parts of code that deal with the safety supervisor, and the core integration aspects is currently closed. We intend for the development of this platform to continue, and for the source code (after obscuring the interfaces we used for our industrial partners) to be released so that others can benefit from AD-EYE as well.

## Acknowledgements

The authors gratefully acknowledge the following projects and agencies for financial support: FFI, Vehicle Strategic Research and Innovation and Vinnova through the ARCHER project (No 2014-06260), H2020 - ECSEL – AutoDrive (Grant Agreement number: 737469), and H2020 - ECSEL PRYSTINE (Grant agreement number 783190).